\documentclass[10pt,journal,compsoc]{IEEEtran}
\ifCLASSOPTIONcompsoc
  \usepackage[nocompress]{cite}
\else
  \usepackage{cite}
\fi
\ifCLASSINFOpdf
\else
\fi
\usepackage{multirow}
\usepackage{algorithm}
\usepackage{algpseudocode}
\usepackage{amsmath}
\usepackage{mathtools}
\usepackage{multirow}
\usepackage{verbatim}
\usepackage[section]{placeins}
\usepackage{float}
\usepackage{amssymb}
\usepackage{subfigure}
\usepackage{graphicx}
\usepackage{url}
\usepackage{hyperref}       
\usepackage{url}            
\usepackage{booktabs}       
\usepackage{amsfonts}       
\usepackage{nicefrac}       
\usepackage{microtype}      
\usepackage{xcolor}         
\usepackage{lineno}

\begin{document}
%
\title{Exhaustive Exploitation of Nature-inspired Computation for Cancer Screening in an Ensemble Manner}
%
%
%
%

\author{Xubin~Wang,
        Yunhe~Wang$^{*}$,
        Zhiqiang~Ma,
        Ka-Chun~Wong,
        and~Xiangtao~Li$^{*}$
\thanks{Xubin Wang and Xiangtao Li are with the School
of Artificial Intelligence, Jilin University, Changchun, Jilin 130012, China, ( e-mail:  wangxb19@mails.jlu.edu.cn; lixt314@jlu.edu.cn). Corresponding authors ($^*$): lixt314@jlu.edu.cn; wangyh082@hebut.edu.cn}
\thanks{Yunhe Wang is with the School of Artificial Intelligence, Hebei University of Technology, Tianjin, China, (e-mail: wangyh082@hebut.edu.cn).}
\thanks{Zhiqiang Ma is with the School of Information Science and Technology, Northeast Normal University, Changchun, Jilin 130117, China, (e-mail: mazq@nenu.edu.cn).}
\thanks{Ka-Chun Wong is with the Department of Computer Science, City University of Hong Kong, Hong Kong SAR, (e-mail: kc.w@cityu.edu.hk).}

}

%
%

\markboth{Journal of \LaTeX\ Class Files,~Vol.~14, No.~8, August~2015}%
{Shell \MakeLowercase{\textit{et al.}}: Bare Demo of IEEEtran.cls for Computer Society Journals}
%



\IEEEtitleabstractindextext{%
\begin{abstract}
Accurate screening of cancer types is crucial for effective cancer detection and precise treatment selection. However, the association between gene expression profiles and tumors is often limited to a small number of biomarker genes. While computational methods using nature-inspired algorithms have shown promise in selecting predictive genes, existing techniques are limited by inefficient search and poor generalization across diverse datasets. This study presents a framework termed Evolutionary Optimized Diverse Ensemble Learning (EODE) to improve ensemble learning for cancer classification from gene expression data. The EODE methodology combines an intelligent grey wolf optimization algorithm for selective feature space reduction, guided random injection modeling for ensemble diversity enhancement, and subset model optimization for synergistic classifier combinations. Extensive experiments were conducted across 35 gene expression benchmark datasets encompassing varied cancer types. Results demonstrated that EODE obtained significantly improved screening accuracy over individual and conventionally aggregated models. The integrated optimization of advanced feature selection, directed specialized modeling, and cooperative classifier ensembles helps address key challenges in current nature-inspired approaches. This provides an effective framework for robust and generalized ensemble learning with gene expression biomarkers. Specifically, we have opened EODE source code on Github at \url{https://github.com/wangxb96/EODE}. 
\end{abstract}

}

\maketitle

\IEEEdisplaynontitleabstractindextext

%
\IEEEpeerreviewmaketitle

\IEEEraisesectionheading{\section{Introduction}\label{sec:introduction}}

\IEEEPARstart{C}{ancer} has become one of the leading causes of mortality worldwide, resulting in over 10 million deaths in 2020 alone \cite{cao2021changing}. The heterogeneity and complexity of various cancer types poses significant challenges for timely and accurate diagnosis, prognosis, and treatment planning \cite{swanson2023patterns, bi2019artificial}. Precision oncology aims to overcome these difficulties by leveraging molecular biomarkers and omics data to guide personalized therapeutic decisions \cite{mateo2022delivering}. In particular, analysis of cancer gene expression data enables identification of discriminative genes and pathways involved in pathogenesis, which can inform diagnostic tests, prognostic indicators, and drug targets \cite{huang2006independent, su2020identification}.

However, several analytical difficulties impose barriers to identifying robust molecular biomarkers from gene expression data. Small sample sizes coupled with extremely high dimensionality and sparsity of the data make computational analysis statistically underpowered \cite{qu2021improving}. Technical noise, batch effects, tumor heterogeneity, and variability between patients also confound analyses \cite{parker2014preserving, schmidt2018ontology}. Effective and robust computational methods are therefore urgently needed to overcome these challenges and accurately detect differentially expressed genes from such complex high-dimensional datasets across diverse cancer types. This can support development of gene expression-based biomarkers for precision oncology applications.

A variety of computational approaches have been applied for cancer gene expression analysis and biomarker identification, including machine learning, deep learning, and nature-inspired optimization algorithms \cite{jin2021ecmarker, he2020integrating, gao2022exploring}. In particular, swarm intelligence and evolutionary algorithms like particle swarm optimization (PSO) \cite{wang2022feature}, ant colony optimization (ACO) \cite{nemati2009novel}, genetic algorithms \cite{maleki2021k}, and enhanced optimizer variants \cite{de2020binary,dhiman2021bepo,hammouri2020improved,neggaz2020efficient} have shown promise. While achieving promising results, further improvements in accuracy, robustness, and generalization ability are still possible. A key limitation is that most methods rely on a single learner algorithm, which makes it difficult to determine the universally optimal learner across diverse cancer types and datasets. Different algorithms have distinct strengths and weaknesses, so their performance varies. Relying on just one also reduces robustness.

Ensemble learning methods which combine multiple diverse base learner models can help address these pitfalls \cite{pratama2018evolving}. Strategies like bagging \cite{breiman1996bagging} and boosting \cite{freund1997decision} train multiple base models on randomized or reweighted data versions, then aggregate predictions to reduce variance and bias. Such ensembles have proven effective for tasks ranging from cancer subtype classification \cite{shen2009integrative, cao2018lnclocator} to drug response modeling \cite{su2020meta}. However, naively combining all base learner models can limit diversity, leading to redundant representations and suboptimal performance \cite{brown2005diversity}. Recent studies have explored intelligent optimizer-guided selection of ensemble subsets to promote specialization and synergy among members \cite{jan2020novel,nguyen2020ensemble,chen2014applying,das2020relevant,li2019single}. For instance, genetic algorithms have been applied to search the space of model combinations, selecting only classifiers that maximize validation accuracy through cooperative interactions \cite{zhu2022genetic}. While showing promise, these approaches generally utilize the full, high-dimensional feature space, which can retain irrelevant variables that confuse models and constrain diversity. Advanced feature selection is needed to derive maximally informative biomarker subsets tailored for ensemble learning \cite{chandrashekar2014survey}. Furthermore, diversity enhancement techniques like bagging and boosting are insufficient to fully overcome representation redundancies during model training \cite{kuncheva2003measures}. Novel forms of controlled randomness injection could better promote specialization by guiding different models to focus on distinct explanatory data facets \cite{zhang2006ensemble, jan2020novel}. Overall there remains great opportunity to advance ensemble classifier performance by integrating intelligent feature selection, guided diversity induction, and metaheuristic optimization of cooperative model combinations \cite{brown2005diversity, rokach2010ensemble}. This can further evolve the state-of-the-art in ensemble methods for precision medicine applications. 

In this work, we propose a novel nature-inspired feature selection algorithm, optimized ensemble classifier, and diversity-enhancing ensemble strategy by integrating the grey wolf optimizer (GWO). Our approach, called Evolutionary Optimized Diverse Ensemble learning (EODE), synergistically combines GWO-based wrapper feature selection, diversity injection via randomized model training, and evolutionary optimization for constructing optimal ensemble classifiers. Specifically, GWO efficiently searches the high-dimensional gene expression space to identify an informative subset of discriminative features for cancer diagnosis. Multiple diverse base classifiers (e.g. SVM, KNN) are trained on these selected features while introducing randomness to increase diversity. Finally, GWO optimizes selection and integration of ensemble members to maximize performance on validation data. EODE enhances generalization ability by leveraging GWO's feature selection, controlled randomness injection, and metaheuristic ensemble optimization. We evaluate EODE on cancer gene expression datasets for tasks including subtype classification, outcome prediction, and the size of feature subset. Results demonstrate EODE significantly improves accuracy and robustness over 23 state-of-the-art methods on 35 cancer gene expression datasets. The integrated strategy advances biomarker discovery and precision oncology by evolving high-performance diverse ensemble classifiers. The main steps of the EODE approach are as follows:

\begin{enumerate}
\item \textit{Base classifiers}: The diversity among the base classifiers is crucial to the effectiveness of the ensemble. The base classifiers can be any suitable classification algorithms, such as decision trees, support vector machines, or neural networks. In this study, six base classifiers including Discriminant Analysis (DISCR), Decision Tree (DT), K-Nearest Neighbor (KNN), Artificial Neural Networks (ANN), Support Vector Machine (SVM), and Naive Bayes (NB) are used.

\item \textit{Classifier selection}: To mitigate the high computational cost associated with using ensemble methods in the feature selection training process, all base classifiers are initially trained with five-fold cross validation using the original training data. The best-performing base classifier is then selected to participate in the feature selection stage. This approach ensures that appropriate learners are involved in training for different datasets to a certain extent.

\item \textit{Feature selection}: GWO is employed to search for an optimal subset of genes that are most relevant to cancer diagnosis. The fitness function is designed to evaluate the quality of each feature subset based on classification performance and the size of feature subset. GWO optimizes the feature subset by iteratively updating the positions of grey wolves based on their fitness values. 

\item \textit{Ensemble diversity enhancement}: To increase the diversity of ensemble, the techniques such as bagging, boosting, or random subspace method can be employed. Here, we generate multiple random subspaces through K-means clustering to increase the diversity of the ensemble. We use these data clusters to train base classifiers, resulting in a pool of models. 

\item \textit{Model pool optimization}: In the model pool, directly fusing all models can lead to lower inference efficiency, and the presence of some low-quality models may degrade the overall performance. Therefore, before final model evaluation, we optimize the model pool. We first performed pre-optimization, discarding models that performed below average on the validation set. For the remaining models, we further optimized using the GWO algorithm to select the possible optimal combination of models.
 
\item \textit{Evaluation and validation}: The performance of the EODE model is evaluated using appropriate metrics such as accuracy, average performance, and the size of the feature subset. The predictions of the selected models are combined using plurality voting. The combined predictions provide the final classification result. Moreover, cross-validation and independent validation datasets are used to assess the generalization ability of the model. 
\end{enumerate}

\section{Methods}
\label{headings}
\subsection{Methodology Overview of EODE}
In this study, we present a novel nature-inspired method called EODE for rapid identification of biomarker genes for multiple cancer types in multiple cancer gene expression datasets. A schematic overview of the algorithm is provided in Figure 1. The original input gene expression data $\mathcal{D}_{or} = \{(x_1,y_1),...,(x_n,y_n)\}$ is considered, where $x_i = (x_{i,1},x_{i,2},...,x_{i,dim})$ represents a sample with $dim$ genes, $y$ belongs to the set $\{1, 2, ..., c\}$ indicating the consensus molecular subtypes, and $n$ is the total number of samples. 

\begin{figure*}[htb!]
\centering
\includegraphics[scale = 0.49]{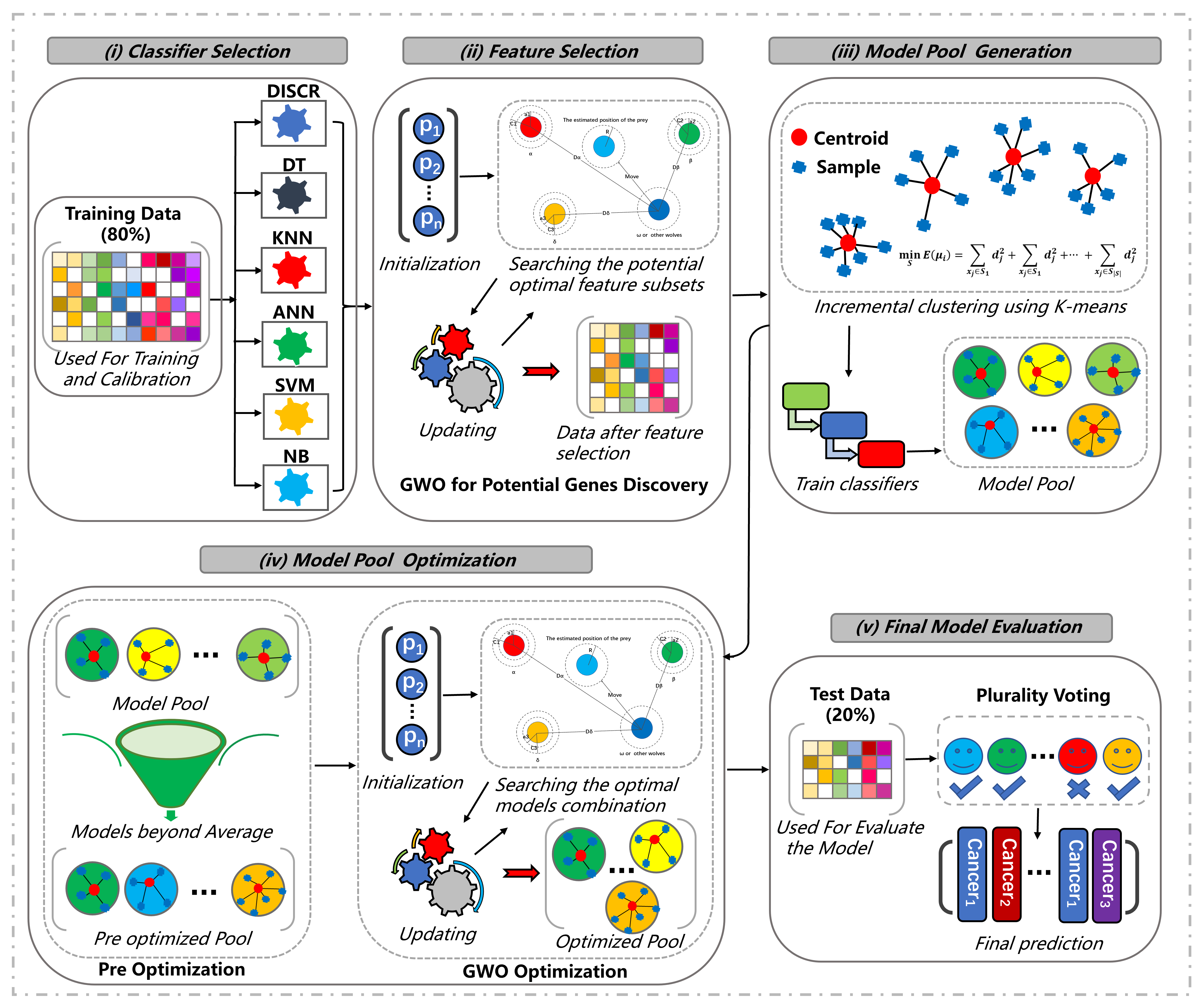}
\caption{Overview of the proposed EODE algorithm: In the GWO feature selection phase, the original cancer gene expression training data is utilized to train all base classifiers, and the classifier with the highest performance is selected as the evaluation classifier. The processed data is then optimized to construct an ensemble model. Specifically, the training data is incrementally clustered using the K-means method to form subspace clusters. These clusters are used to train individual base classifiers, which are then added to the model pool. Any classifiers in the pool with below-average performance are filtered out. Next, the GWO is applied to optimize the classifier pool and determine the best possible ensemble combination. Finally, the optimized ensemble model is evaluated on the independent test dataset using a plurality voting strategy to generate the final cancer type predictions.}
\label{fig:graph}
\end{figure*}

In the feature selection step, we employ the GWO to extract relevant biomarker genes after training our model on the training gene expression matrix $\mathcal{D}_{tr}$. Each base classifier from the pool $\mathcal{B}$ (including Discriminant Analysis (DISCR), Decision Tree (DT), K-Nearest Neighbor (KNN), Artificial Neural Networks (ANN), Support Vector Machine (SVM), and Naive Bayes (NB)) is initially trained using the input data. The best-performing classifier is then chosen as the evaluation classifier for feature selection.

The processed data is subsequently utilized to train and optimize a diverse ensemble model. Specifically, the data undergoes five-fold cross-validation to construct the final model $\Psi$. Initially, the data is partitioned into progressive subspaces using the K-means method to form clusters. These clusters are then utilized to train base classifiers, which are subsequently incorporated into the model pool. Models in the pool with below-average performance are filtered out. After that, the GWO approach is applied to optimize the model pool and identify the best possible combination. Finally, the model $\Psi$ is evaluated on the test data using a plurality voting strategy. The overall framework of EODE is summarized in Algorithm 1.

{
\linespread{1.0} \selectfont
\begin{algorithm}[!htb]
  \caption{Pseudo Code of EODE Algorithm}
  \label{alg:Framwork}
  \begin{algorithmic}[1]
    \Require Training Data: $\mathcal{D}_{tr} =\{(x_{tr,1},y_{tr,1}), ..., (x_{tr,n},y_{tr,n})\}$, $x \in R^d$, $y \in \{1,2,...,c\}$, Test Data:  $\mathcal{D}_{te} =\{(x_{te,1},y_{te,1}),..., (x_{te,n},y_{te,n})\}$, a set of base classifiers $\mathcal{B}$, upper bounds of clustering $K$, population of GWO $\vec{X}$, the feature selection function $f_1$, the classifiers optimization function $f_2$
    \State Use training data $\mathcal{D}_{tr}$ to train each base classifier in $\mathcal{B}$
    \State The classifier $b$ with the best performance is selected for feature selection  
    \State Initialize a population of $|\vec{X}|$ individuals
    \While{$t <$ max iterations $T$}
    \State $\vec{X} \xleftarrow{} $  Use \textit{Algorithm 2} to do $f_1(\vec{X})$
    \State $\vec{X_i} \xleftarrow{}$ best individual
    \State $t$++;
    \EndWhile
    \State $bf \xleftarrow{}$ best features selected by wolf $\vec{X_i}$ 
    \State $fnum \xleftarrow{}$ the number of features selected by wolf $\vec{X_i}$
    \State $\mathcal{D}_{tr} \xleftarrow{} \mathcal{D}_{tr}(bf)$
    \State $\mathcal{D}_{te} \xleftarrow{} \mathcal{D}_{te}(bf)$
    \State $\mathcal{D}_{tr} =\mathcal{D}_{tr,1} \cup ...\cup \mathcal{D}_{tr,5}, \mathcal{D}_{tr,i} \cap \mathcal{D}_{tr,j} = \emptyset (i \neq j) $
    \For{each $\mathcal{D}_{tr,i}$ in $\mathcal{D}_{tr}$}
       \State $\mathcal{D}_{tr}^{-i} = \mathcal{D}_{tr} - \mathcal{D}_{tr,i}$
       \For{$k = 1 \xrightarrow{} K$}  
       \State $C^S \xleftarrow{}$ partition $\mathcal{D}_{tr}^{-i}$ into $k$ clusters
       \State $S = S + 1$
       \EndFor
       \For{each $mp_i$ in $\mathcal{MP}$}
           \State $Acc(i) \xleftarrow{}$ calculate each $mp_i$'s validation accuracy on $D_{tr,i}$ 
       \EndFor
       \State $\mathcal{MP} \xleftarrow{}$ ($mp_i$ if $Acc(mp_i) >$ mean($Acc$))
       \State Initialize a population of $|\vec{X}|$ individuals
       \While{$t <$ max iterations $T$}
       \State $\vec{X} \xleftarrow{}$ Use \textit{Algorithm 2} to do $f_2(\vec{X})$
       \State $\vec{X_i} \xleftarrow{}$ best individual
       \State $t$++;
       \EndWhile
       \State $\psi \xleftarrow{}$ best models in $\mathcal{MP}$ selected by wolf $\vec{X_i}$
       \State Optimized classifier $\Psi \xleftarrow{} \Psi + \psi$
    \EndFor
    \State $testAcc \xleftarrow{}$ classify samples of $\mathcal{D}_{te}$ by $\Psi$
    \State \textit{Output:} The optimized ensemble classifier $\Psi$, the number of selected features $fnum$ and the test accuracy $testAcc$
  \end{algorithmic}
\end{algorithm}
}

\subsection{Nature-inspired Feature Selection}
Considering a training cancer gene expression data $\mathcal{D}_{tr} = \{(x_1,y_1),...,(x_n,y_n)\}$, where $x_i = (x_{i,1},x_{i,2},...,x_{i,dim})$ represents the feature vector and $dim$ denotes the number of features, $y$ belongs to the set $\{1,2,..,c\}$ representing the class, and $n$ is the number of samples. It is important to note that the high-dimensional nature of the gene expression data may include many irrelevant genes, which can negatively impact identification accuracy while increasing computational time \cite{qu2021improving}. Therefore, performing feature selection is crucial to preprocess the data effectively.

The Grey Wolf Optimizer (GWO), initially proposed by Mirjalili \cite{mirjalili2014grey}, is a swarm intelligence algorithm inspired by the social hierarchy and hunting behavior of grey wolves in nature. GWO offers advantages such as good convergence, minimal parameter tuning, and ease of implementation \cite{faris2018grey}. The core concept of GWO revolves around three primary predation behaviors: encircling prey, hunting, and attacking prey, which are performed based on the social hierarchy among the wolves. The social hierarchy in GWO consists of four levels: $\alpha$, $\beta$, $\delta$, and $\omega$, with $\alpha$ being the dominant wolf, followed by $\beta$ and $\delta$, while the remaining wolves are labeled as $\omega$. Wolves at higher ranks exert dominance over those at lower ranks, and $\alpha$, $\beta$, and $\delta$ play key roles in the algorithm, with $\alpha$ being the wolf king and $\beta$ and $\delta$ serving as potential successors. The $\alpha$ wolf represents the fittest solution and guides the pack towards promising search areas. The second and third best fit solutions are modeled as $\beta$ and $\delta$ wolves, respectively. The $\omega$ wolves represent the remaining weaker candidate solutions that follow the guidance of the $\alpha$, $\beta$ and $\delta$ wolves. During optimization, the candidate solutions iteratively update their positions towards the best three solutions until convergence upon the global optimal value. Specifically, a schematic representation of GWO is depicted in Fig. 2.

Building upon these foundations, we propose a nature-inspired feature selection method based on GWO, which comprises six essential components: classifier selection, population initialization, encircling prey phase, hunting phase, attacking phase, and feature selection objective function.

\subsubsection{Classifier Selection}
To evaluate the feature selection results, we consider six base classifiers in a classifier pool $\mathcal{B}$: Discriminant Analysis (DISCR), Decision Tree (DT), K-Nearest Neighbor (KNN), Artificial Neural Networks (ANNs), Support Vector Machine (SVM), and Naive Bayes (NB). However, incorporating all these classifiers into the ensemble method during the feature selection phase would be computationally expensive. Therefore, we adopt a pre-training approach to select the best-performing classifier from the pool $\mathcal{B}$. The cancer gene expression data $\mathcal{D}$ is subjected to five-fold cross-validation on each base classifier, and the classifier with the highest performance is chosen as the evaluation classifier for the feature selection phase. This approach allows us to efficiently select the most suitable classifier for the subsequent feature selection process.

\subsubsection{Population Initialization}
In the beginning, the population $\vec{X}$ is randomly created and represented as real numbers. Each individual, denoted as $\vec{X_i}$, is a set of genes: $\vec{X_i} = \{ g_1, g_2, ..., g_{dim} \}$, where $g_{dim}$ represents the $dim$th gene and $dim$ is the total number of genes. 

To convert these real numbers into a binary form, we use a threshold value $\theta$. If a feature value ($g_n$) is greater than or equal to $\theta$, it is set to 1, indicating that the corresponding feature is selected. On the other hand, if $g_n$ is less than $\theta$, it is set to 0, indicating that the feature is not selected. The process can be as follows:

\begin{equation}
     g_n =\begin{dcases}
1,&g_n \geq \theta\\
0,&g_n < \theta
\end{dcases}.
\end{equation}

After that, the position of each individual is represented by a binary (0/1) string.  

\subsubsection{Encircling Prey Phase}
The "encircling prey" behavior is a strategy employed by the grey wolf pack to search for feature subsets. This behavior is mathematically modeled to simulate how the grey wolf gradually approaches its prey and surrounds it. The distance ($\vec{D}$) between the grey wolf and the prey is determined by the equation:
\begin{equation}
    \vec{D} = |2 \cdot r_2 \cdot \vec{X_p(t)} - \vec{X_i(t)}|,
\end{equation}
where $\vec{D}$ represents the distance between them. During the search process, the current iteration is denoted by $t$, and $\vec{X_p(t)}$ and $\vec{X_i(t)}$ represent the position vectors of the prey and the grey wolf, respectively.

To update the position of the grey wolf, we utilize the formula $ \vec{X_i(t+1)} = \vec{X_p(t)} - (2\vec{a} \cdot r_1 - \vec{a}) \cdot \vec{D}$. Here, $\vec{a}$ is the convergence factor that decreases linearly from 2 to 0 as the iterations progress. The convergence factor is calculated as $\vec{a} = 2 - 2t$/$max_t$, where $t$ represents the current iteration, and $max_t$ is the maximum number of iterations defined for the search process. Additionally, $r_1$ and $r_2$ are random numbers between 0 and 1.

By applying this position update formula, the grey wolf adjusts its position towards the prey. The term $(2\vec{a} \cdot r_1 - \vec{a})$ determines the magnitude and direction of the movement, while the distance $\vec{D}$ guides the grey wolf's movement in narrowing the gap with the prey. The process continues iteratively until the desired maximum number of iterations is reached ($max_t$). Ultimately, the grey wolf is expected to encircle the prey, indicating the discovery of a promising feature subset.

{
\linespread{1.0} \selectfont
\begin{algorithm}[H]
\caption{Pseudo Code of Grey Wolf Optimizer (GWO)}
\begin{algorithmic}[1]
\State Initialize a population $\vec{X}$ of wolves randomly within the solution space
\State Evaluate the fitness of each wolf $\vec{X_i}$ using a fitness function $f$
\State Set the initial values for $\alpha$, $\beta$, and $\delta$ as the wolves with the highest, second highest, and third highest fitness, respectively
\While{t $<$ max iterations $T$}
\For{each wolf $\vec{X_i}$ in the population}
\State Update the position of the wolf based on the positions of $\alpha$, $\beta$, and $\delta$ using the following formulas:
\State $\vec{D}_\alpha = |\vec{C}_1 \cdot \vec{X}_\alpha - \vec{X}_i|$ // Distance from $\alpha$
\State $\vec{D}_\beta = |\vec{C}_2 \cdot \vec{X}_\beta - \vec{X}_i|$ // Distance from $\beta$
\State $\vec{D}_\delta = |\vec{C}_3 \cdot \vec{X}_\delta - \vec{X}_i|$ // Distance from $\delta$
\State $\vec{X}' = \vec{X}_\alpha - A_1 \cdot \vec{D}_\alpha$ // Encircling $\alpha$
\State $\vec{Y}' = \vec{X}_\beta - A_2 \cdot \vec{D}_\beta$ // Encircling $\beta$
\State $\vec{Z}' = \vec{X}_\delta - A_3 \cdot \vec{D}_\delta$ // Encircling $\delta$
\State Update the position of the wolf using:
\State $\vec{X_i(t+1)} = (\vec{X}' + \vec{Y}' + \vec{Z}') / 3$
\State Apply boundary constraints to ensure the new position is within the solution space
\EndFor
\State Update the fitness of each wolf $\vec{X_i}$ using a fitness function $f$
\State Update $\alpha$, $\beta$, and $\delta$ based on the updated fitness values
\State $t$++;
\EndWhile
\State \textit{Output:} The position of the $\alpha$ wolf represents the best solution found by the GWO algorithm
\end{algorithmic}
\end{algorithm}
}

\begin{figure}[htb]
\centering
\includegraphics[scale = 0.28]{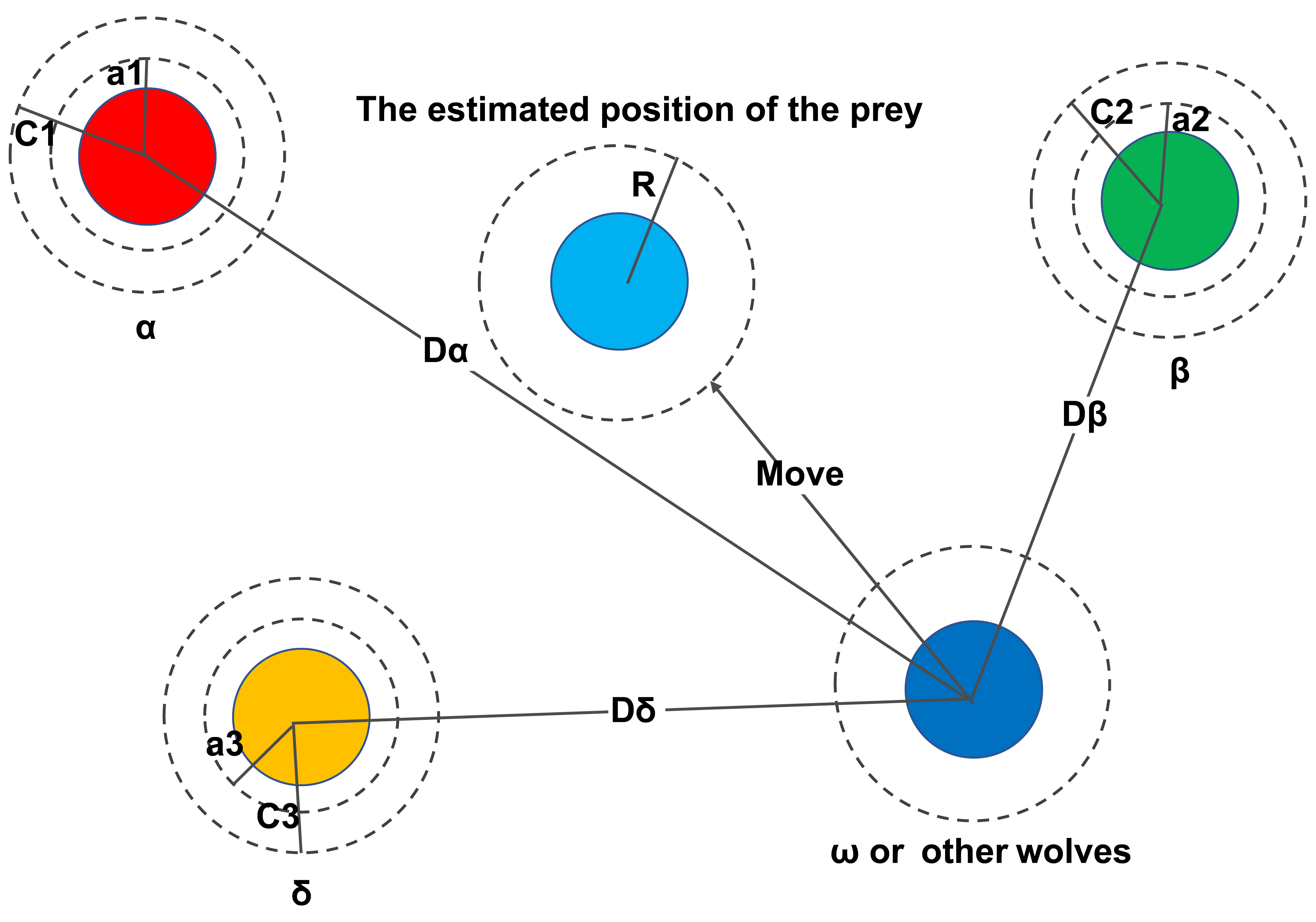}
\caption{The GWO algorithm is illustrated in a schematic representation, highlighting the process of updating the positions of the wolves. Initially, the positions of the wolves are randomly initialized within the solution space. The fitness of each wolf is evaluated based on a fitness function. In each iteration, the positions of the wolves are updated using mathematical formulas that consider the social hierarchy, with the $\alpha$ wolf having the greatest influence. The update process involves attracting other wolves towards the positions of the $\alpha$, $\beta$, and $\delta$ wolves. This iterative position updating continues until a termination condition is met. Ultimately, the position of the $\alpha$ wolf represents the best solution found by the GWO algorithm.}
\label{fig:graph}
\end{figure}

\subsubsection{Hunting Phase}
Grey wolves possess the ability to identify the general location of their prey and work together to surround it. However, in many unknown situations, they may not have precise knowledge of the exact location of the target. In our study, we simulate the behavior of grey wolves by introducing three key individuals: $\alpha$, $\beta$, and $\delta$. These individuals help guide the entire wolf pack in surrounding the prey and searching for the optimal solution.

To track the position of the prey, each individual grey wolf calculates its distance to the prey using the following equations:
\begin{equation}
    \vec{D}_\alpha = |\vec{C}_1 \cdot \vec{X}_\alpha - \vec{X}_i|,
\end{equation}
\begin{equation}
    \vec{D}_\beta = |\vec{C}_2 \cdot \vec{X}_\beta - \vec{X}_i|,
\end{equation}
\begin{equation}
    \vec{D}_\delta = |\vec{C}_3 \cdot \vec{X}_\delta - \vec{X}_i|.
\end{equation}

Here, $\vec{D_{\alpha}}$, $\vec{D_{\beta}}$, and $\vec{D_{\delta}}$ represent the distances between the grey wolves $\alpha$, $\beta$, $\delta$ and the prey, respectively. $\vec{X_{\alpha}}$, $\vec{X_{\beta}}$, and $\vec{X_{\delta}}$ denote the positions of $\alpha$, $\beta$, and $\delta$, while $\vec{X_i}$ represents the current position of the grey wolf. Additionally, $\vec{C_1}$, $\vec{C_2}$, and $\vec{C_3}$ are random vectors used to calculate these distances.

Each grey wolf updates its position based on these distance calculations:
\begin{equation}
    \vec{X}' = \vec{X}_\alpha - A_1 \cdot \vec{D}_\alpha,
\end{equation}
\begin{equation}
    \vec{Y}' = \vec{X}_\beta - A_2 \cdot \vec{D}_\beta,
\end{equation}
\begin{equation}
    \vec{Z}' = \vec{X}_\delta - A_3 \cdot \vec{D}_\delta.
\end{equation}

Here, $\vec{X}'$, $\vec{Y}'$, and $\vec{Z}'$ represent the new positions of the grey wolves moving towards $\alpha$, $\beta$, and $\delta$, respectively. The constants $A_1$, $A_2$, and $A_3$ control the magnitude of the movement towards the prey.

Finally, the position of the grey wolf at the next time step $\vec{X_i(t+1)}$ is determined as the average of the positions $\vec{X}'$, $\vec{Y}'$, and $\vec{Z}'$:
\begin{equation}
    \vec{X_i(t+1)} = \frac{\vec{X}' + \vec{Y}' + \vec{Z}'}{3}.
\end{equation}

In this way, the entire wolf pack moves together towards the positions of $\alpha$, $\beta$, and $\delta$, and the new position of each individual is updated accordingly.

\subsubsection{Attacking Phase}
The final stage of the hunting process is the attack, during which the grey wolves aim to capture their prey and obtain the optimal solution. This phase involves adjusting certain parameters to strike a balance between global exploration and local exploitation.

To achieve this balance, two key parameters are considered: $a$ and $A$. The value of $a$ is progressively decreased from 2 to 0 in a linear manner. Simultaneously, the range of fluctuations in $A$ is reduced. The parameter $A$ takes on values within the range $[-a, a]$. The behavior of the grey wolves is influenced by the magnitude of $A$. When the absolute value of $A$ is greater than 1, the grey wolves tend to spread out across different areas, enabling a global search for prey. Conversely, when the absolute value of $A$ is less than 1, the grey wolves exhibit a more focused, local search.

In addition to these parameters, the influence of the grey wolves' positions on the prey is governed by a random weight, denoted as $C$. This weight, which ranges between 0 and 2, determines the random influence of the grey wolf's location on the prey. A value of $C$ greater than 1 indicates a higher weight, emphasizing the significance of the grey wolf's position in guiding the search. Conversely, a value of $C$ less than 1 assigns a lower weight, reducing the impact of the grey wolf's location. This random weight, $C$, helps prevent the algorithm from converging too early and becoming trapped in a local optimum.

By dynamically adjusting the values of $a$, $A$, and $C$ during the attacking phase, the grey wolves strike a balance between exploration and exploitation, allowing them to efficiently search for and capture the optimal solution while avoiding premature convergence and local optima.

\subsubsection{Feature Selection Objective Function}
During each iteration of the GWO algorithm, the classification label for each candidate solution $\vec{X_i}$ is predicted using the evaluation classifier selected from the classifier selection phase. Specifically, the evaluation classifier is initially trained on the original training gene expression dataset $\mathcal{D}_{tr}$ with all features using five-fold cross-validation. For each $\vec{X_i}$ containing a subset of selected features, the evaluation classifier generates predicted labels $y_i'$ by classifying the corresponding data points from $\mathcal{D}_{tr}$ using only the selected features in $\vec{X_i}$. The performance of $y_i'$ on $\mathcal{D}_{tr}$ determines the fitness value assigned to solution $\vec{X_i}$. This allows the GWO algorithm to determine the $\alpha$, $\beta$, and $\delta$ solutions representing the current best feature subsets for classification.

In the feature selection stage, the primary objective is to identify and select relevant features while filtering out redundant ones for subsequent identification purposes in cancer gene expression data. Traditional studies often focus solely on classification accuracy, disregarding the resource costs associated with redundant features. In our study, we address this limitation by considering both classification accuracy and the size of the feature subsets as part of our feature selection objective function \cite{xue2012particle}.

The objective function, denoted as $f_1$, is defined as follows:
\begin{equation}
f_1 = \alpha * \text{error} + \beta * \frac{f_{\text{num}}}{\text{dim}}.
\end{equation}

Here, $f_{\text{num}}$ represents the number of selected features during the evolutionary process, and $\text{dim}$ represents the total number of features in the dataset. To strike a balance between the two objectives, we introduce weight coefficients to control their relative importance. In our study, we assign a weight of 0.9 to $\alpha$ to emphasize the significance of classification accuracy, while $\beta$ is set to 0.1 to underscore the importance of the feature subset size. These weight coefficients were determined based on the findings in the reference \cite{wang2007feature}, where classification accuracy was identified as the primary objective.

The classification error ($\text{error}$) is a key component of the objective function. It is calculated as the difference between 1 and the accuracy ($\text{acc}$), which is defined as:
\begin{equation}
\text{error} = 1 - \text{acc},
\end{equation}
\begin{equation}
\text{acc} = \frac{\sum_{s=1}^n{I(y'_s,y_s)}}{n}.
\end{equation}

In the above equations, $n$ represents the total number of instances, $y'_s$ represents the predicted class label for instance $s$, and $y_s$ represents the true class label for instance $s$. The function $I(y'_s,y_s)$ evaluates to 1 if the predicted and true class labels match, and 0 otherwise.

\subsection{Nature-inspired Diverse Ensemble Learning}
In this section, we propose a novel nature-inspired diverse ensemble learning method to improve the performance of cancer identification using selected features obtained through nature-inspired feature selection. Our method comprises diverse subspace generation, model pool generation, and model pool optimization.

\subsubsection{Diverse Subspace Generation} 
Given the gene expression data after feature selection, denoted as $\mathcal{D}_{tr}' = \{(x_1,y_1),...,(x_m,y_m)\}$, where $x_i =(x_{i,1}, x_{i,2}, ..., x_{i,dim})$ represents the feature vector with $dim$ denoting the number of features, $y\in \{1,2,...,c\}$ represents the classification label, and $m$ represents the number of input samples, we employ the K-means method \cite{mackay2003information} to cluster the input cancer gene expression data into multiple clusters. The clustering process is performed iteratively from 1 to $t$, generating $K$ clusters in each iteration. Here, $t$ denotes the total number of iterations. The clusters are obtained by minimizing the following function:
\begin{equation}
\mathop{argmin}\limits_S \sum_{i=1}^K\sum_{x \in S_i}{||x - \mu_i||^2},
\end{equation}
where $x$ represents the feature vector and $\mu_i$ is the centroid of cluster $S_i$. This clustering process generates a set of diverse subspaces composed of all the obtained clusters.

\subsubsection{Model Pool Generation}
Each cluster in the diverse subspace is used to train six classifiers (DISCR, DT, KNN, ANN, SVM, and NB) to create a model pool. The base classifiers used in this step are independent. The resulting models are then added to the base model pool $\mathcal{MP}$. The base model pool $\mathcal{MP}$ consists of $l * |\mathcal{B}|$ models, where $l$ represents the number of clusters and $|\mathcal{B}|$ represents the number of base classifiers. Finally, we employ nature-inspired optimization techniques to refine the base models in the ensemble. Here, any combination of classifiers can be utilized.

\subsubsection{Model Pool Optimization}
After obtaining the diverse base model pool $\mathcal{MP}$, we propose a pre-optimization step to refine $\mathcal{MP}$ by removing models with below-average performance. Subsequently, we incorporate a nature-inspired optimization method, namely GWO, to further optimize the pre-optimized base model pool $\mathcal{MP}$.

\textit{Population Initialization:} The population is randomly initialized, and each individual is represented as follows:
\begin{equation}
\Vec{X_i} = \{mp_1, mp_2,..., mp_r\}.
\end{equation}

Here, $mp_r$ represents a classifier in the model pool $\mathcal{MP}$, and $r$ is the total number of models in $\mathcal{MP}$. Similar to nature-inspired feature selection, the selection or non-selection of models is indicated by binary values. "1" indicates that a model is selected, while "0" indicates that the model is not selected. To convert the continuous search space of GWO into a binary search space, we introduce a threshold $\theta$. The conversion from a continuous position to discrete binary values is defined as follows:
\begin{equation}
mp_r =\begin{dcases}
1,&mp_r \geq \theta\\
0,&mp_r < \theta
\end{dcases}.
\end{equation}

\textit{Nature-inspired Optimization Process}: In this phase, our aim is to discover optimal model subsets by optimizing the base model pool $\mathcal{MP}$. The population is used to explore optimal model subsets in the encircling phase, identify potential optimal solutions in the hunting phase, and ultimately obtain the optimal solution in the attacking phase.

\textit{Ensemble Optimizing Objective Function}: Our objective is to achieve the highest identification performance with the smallest ensemble size. After clustering the data following feature selection and training the base classifiers to create a model pool, we aim to optimize the model pool to obtain the optimal ensemble model with the smallest size. The optimized model ensemble is then evaluated using the test data. The objective function in the model pool optimization stage, denoted as $f_2$, is defined as follows:
\begin{equation}
    f_2 = \alpha * \text{error} + \beta * \frac{|\psi|}{r}.
\end{equation}

Here, $\text{error}$ represents the identification error rate described in Equation (11), $|\psi|$ is the total number of selected models, and $r$ is the number of models in $\mathcal{MP}$. The settings of $\alpha$ and $\beta$ are identical to those in section 2.2.6, with $\alpha$ accounting for 90\% of the importance and $\beta$ for 10\%. 

However, unlike in section 2.2.6, where the predicted label $y'_s$ is predicted by a single classifier, we consider the ensemble of multiple models. We employ a plurality voting method to combine the predictions of multiple models, which has been proven to be a simple and effective ensemble fusion technique in many studies \cite{meir2010convergence} \cite{meir2015plurality}.

\subsubsection{Ensemble Classifier Prediction}
During the training process, we obtain multiple models $\psi$ to represent the model $\Psi$. The model $\Psi$ is used to generate an ensemble, and all models in $\Psi$ are utilized to predict the test set. The predicted class labels $y'_s$ from all the models in the model $\Psi$ are fused using the plurality voting method. The identification accuracy can be calculated using Equation (12).

\subsection{Time Complexity Analysis}
Here, we analyze the time complexity of our proposed EODE algorithm. The detailed analysis is outlined as follows: 
\begin{itemize}
\item \textit{Feature Selection:} The time complexity of the feature selection process depends on the algorithm used. Since we used GWO for feature selection, the time complexity is typically $O(T \times P \times F \times C)$, where $T$ is the number of generations, $P$ is the population size, $F$ is the number of features, and $C$ is the complexity of the fitness evaluation function. Generally, the feature selection process has a polynomial time complexity.

\item \textit{Diverse Subspace Generation:} The time complexity of the diverse subspace generation mainly depends on the clustering algorithm used. Here, we applied the K-means algorithm, the time complexity is usually $O(K \times N \times I \times d)$, where $K$ is the number of clusters, $N$ is the number of data points, $I$ is the number of iterations, and $d$ is the dimensionality of the data. The diverse subspace generation process has a polynomial time complexity.

\item \textit{Model Pool Generation:} The model pool generation involves training multiple base classifiers on each cluster. The time complexity depends on the complexity of the base classifiers and the number of clusters. Assuming the time complexity of training a base classifier on a single cluster is $O(N \times F \times C)$, where $N$ is the number of data points, $F$ is the number of selected features, and $C$ is the complexity of the training algorithm, the overall time complexity of model pool generation is $O(L \times N \times F \times C)$, where $L$ is the number of clusters. This process also has a polynomial time complexity.

\item \textit{Model Pool Optimization:} The time complexity of the model pool optimization stage depends on the optimization algorithm used. We employed a nature-inspired optimization algorithm named GWO, the time complexity is typically $O(T \times P \times C)$, where $T$ is the number of generations, $P$ is the population size, and $C$ is the complexity of the fitness evaluation function. Similar to the feature selection process, the model pool optimization stage generally has a polynomial time complexity.

\item \textit{Ensemble Classifier Prediction:} The time complexity of the ensemble classifier prediction is dependent on the number of models in the ensemble and the complexity of combining their predictions. Assuming we have $M$ models in the ensemble and the complexity of combining predictions is $O(M)$, the overall time complexity is $O(M)$. This process has a linear time complexity.
\end{itemize}
In summary, the overall time complexity is:
\text{Overall Time Complexity} = \text{Feature Selection} + \text{Diverse Subspace Generation} + \text{Model Pool Generation} + \text{Model Pool Optimization} + \text{Ensemble Classifier Prediction} = O(T $\times$ P $\times$ F $\times$ C) + O(K $\times$ N $\times$ I $\times$ d) + O(L $\times$ N $\times$ F $\times$ C) + O(T $\times$ P $\times$ C) + O(M)

Since all these time complexities are polynomial, we can express the overall time complexity as the highest-order term in the sum. Therefore, the overall time complexity of the EODE algorithm is: 

\text{Overall Time Complexity} = $\max$\{T $\times$ P $\times$ F $\times$ C, K $\times$ N $\times$ I $\times$ d, L $\times$ N $\times$ F $\times$ C, T $\times$ P $\times$ C, M\}

\section{Implementation}
\label{others}

\subsection{Datasets}
The cancer gene expression datasets were collected from \cite{de2008clustering}, and can be downloaded from the website \url{https://schlieplab.org/Static/Supplements/CompCancer/datasets.htm}. The 35 datasets contain multiple types of cancers with high-dimensional features, exceeding 1000 dimensions, while having relatively small sample sizes (as shown in TABLE 1). This poses the ``Curse of Dimensionality" challenge, necessitating the development of a computational model with high robustness and good generalization capabilities to address the different cancers.

To enable rigorous evaluation, the collected raw datasets have been randomly split into disjoint training and testing sets in a 80:20 ratio prior to conducting experiments. The training sets, comprising 80\% of the data, have been used for model training and hyperparameter tuning. The testing sets, comprising the held-out 20\% of the data, have only been used for final evaluation of the fully trained model's performance. This ensures an unbiased estimate of generalization capability. The precise training/testing splits has been done randomly while preserving class balance in each set. Specifically, the training and testing datasets can be downloaded from the following links: \url{https://github.com/wangxb96/EODE/tree/master/TrainData} and \url{https://github.com/wangxb96/EODE/tree/master/TestData}. For model selection and hyperparameter tuning, k-fold cross-validation (k=5) was utilized during model selection and hyperparameter optimization on the training data only. By segregating the training and testing data, we prevent information leakage and overfitting to the test set. This rigorous methodology allows us to evaluate true generalization error and robustness across multiple cancer types. 

{

\begin{table*}[!htb]
\centering
\caption{35 different gene expression datasets; each dataset showing the tissue type, number of samples, features, and classes.}
\resizebox{178mm}{!}{
\begin{tabular}{| l | l | l | l | l || l | l | l | l | l | }
\hline
\textbf{Dataset} & \textbf{Tissue} & \textbf{Samples}& \textbf{Features} & \textbf{Classes} & \textbf{Dataset} & \textbf{Tissue} & \textbf{Samples}& \textbf{Features} & \textbf{Classes}\\
\hline \hline
Alizadeh-2000-v1	&	Blood	&	42	&	1095	&	2	 &
Alizadeh-2000-v2	&	Blood	&	62	&	2093	&	3	 \\
\hline
Alizadeh-2000-v3	&	Blood	&	62	&	2093	&	4		& Armstrong-2002-v1	&	Blood	&	72	&	1081	&	2  \\
\hline
Armstrong-2002-v2	&	Blood	&	72	&	2194	&	3 & Bhattacharjee-2001	&	Lung	&	203	&	1543	&	5	 \\
\hline
Bittner-2000	&	Skin	&	38	&	2201	&	2 
& Bredel-2005	&	Brain	&	50	&	1739	&	3	\\ 
\hline
Chen-2002	&	Liver	&	179	&	85	&	2	 
& Chowdary-2006	&	Breast, Colon	&	104	&	182	&	2	 \\
\hline
Dyrskjot-2003	&	Bladder	&	40	&	1203	&	3 
& Garber-2001	&	Lung	&	66	&	4553	&	4	 	\\
\hline
Golub-1999-v1	&	Bone Marrow	&	72	&	1877	&	2	 	& Golub-1999-v2	&	Bone Marrow	&	72	&	1877	&	3	 	\\
\hline
Gordon-2002	&	Lung	&	181	&	1626	&	2 & Khan-2001	&	Multi-tissue	&	83	&	1069	&	4	 	\\
\hline
Laiho-2007	&	Colon	&	37	&	2202	&	2	& Lapointe-2004-v1	&	Prostate	&	69	&	1625	&	3	 	\\
\hline
Lapointe-2004-v2	&	Prostate	&	110	&	2496	&	4	& Liang-2005	&	Brain	&	37	&	1411	&	3	 	\\
\hline
Nutt-2003-v1	&	Brain	&	50	&	1377	&	4 	& Nutt-2003-v2	&	Brain	&	28	&	1070	&	2	 	\\
\hline
Nutt-2003-v3	&	Brain	&	22	&	1152	&	2	 & Pomeroy-2002-v1	&	Brain	&	34	&	857	&	2	 	\\
\hline
Pomeroy-2002-v2	&	Brain	&	42	&	1379	&	5	 & Ramaswamy-2001	&	Multi-tissue	&	190	&	1363	&	14 \\	 
\hline
Risinger-2003	&	Endometrium	&	42	&	1771	&	4	 & Shipp-2002-v1	&	Blood	&	77	&	798	&	2 	\\
\hline
Singh-2002	&	Prostate	&	102	&	339	&	2 & Su-2001	&	Multi-tissue	&	174	&	1571	&	10 	\\
\hline
Tomlins-2006-v1	&	Prostate	&	104	&	2315	&	5	 &
Tomlins-2006-v2	&	Prostate	&	92	&	1288	&	4 \\
\hline
West-2001	&	Breast	&	49	&	1198	&	2	 &
Yeoh-2002-v1	&	Bone Marrow	&	248	&	2526	&	2 	\\
\hline
Yeoh-2002-v2	&	Bone Marrow	&	248	&	2526	&	6 & - & - & - & -& -\\
\hline
\end{tabular}}
\end{table*}

}

\subsection{Baselines}
To evaluate the effectiveness of our proposed method, we compared it against several existing classifiers and ensemble algorithms widely used in the literature. Firstly, we compared our model with six base classifiers: DISCR (Discriminant Analysis) \cite{lachenbruch1979discriminant}, DT (Decision Tree) \cite{safavian1991survey}, KNN (K-Nearest Neighbor) \cite{altman1992introduction}, ANN (Artificial Neural Networks) \cite{yegnanarayana2009artificial}, SVM (Support Vector Machine) \cite{noble2006support}, and NB (Naive Bayes) \cite{murphy2006naive}. These classifiers serve as the baseline for performance comparison.

Next, we compared our approach with seven evolutionary algorithms: ACO \cite{dorigo2006ant}, CS \cite{yang2009cuckoo}, DE \cite{das2010differential}, GA \cite{whitley1994genetic}, GWO \cite{mirjalili2014grey}, PSO \cite{kennedy1995particle}, and ABC \cite{karaboga2007powerful}. These algorithms are widely used for optimization problems. Furthermore, we evaluated our approach against four novel ensemble methods: PSOEL \cite{jan2020novel}, EAEL \cite{md2019evolutionary}, FESM \cite{jan2020anovel}, and GA-Bagging-SVM \cite{lin2019accurate}. These methods were selected to demonstrate the effectiveness of our proposed approach in comparison to recent advancements in ensemble learning.

In addition, we compared our ensemble algorithm with six state-of-the-art ensemble classifiers: Random Forests (RF) \cite{breiman2001random}, ADABOOST \cite{freund1997decision}, RUSBOOST \cite{seiffert2009rusboost}, SUBSPACE \cite{ho1998random}, TOTALBOOST \cite{warmuth2006totally}, and LPBOOST \cite{grove1998boosting}. Random Forests is a well-known bagging method \cite{breiman2001random}, while ADABOOST is a popular boosting method \cite{freund1997decision}. RUSBOOST is a random undersampling boosting method designed to address class imbalance \cite{seiffert2009rusboost}. SUBSPACE trains random feature subsets to reduce estimator correlation \cite{ho1998random}. TOTALBOOST and LPBOOST aim to maximize the minimal margin of learned ensembles and have the ability to self-terminate \cite{warmuth2006totally} \cite{grove1998boosting}.

By comparing our method against these diverse algorithms, we aim to showcase its superiority and effectiveness in addressing the cancer gene expression data classification problem. Moreover, we
have opened all computational model for public accessibility at ``\url{https://github.com/wangxb96/EODE/tree/master/ComparisonAlgorithms}''. 

\subsection{Parameter Settings}
Our experiments were conducted on a desktop computer with the following specifications: an Intel(R) Core(TM) i7-10700KF CPU @3.80GHz, 32GB of RAM, and a 64-bit Windows 10 operating system using Matlab 2021a. We utilized six base classifiers, namely DISCR, DT, KNN, ANN, SVM, and NB, to construct the ensemble. The parameters for DISCR, KNN, SVM, and NB are summarized in TABLE II, while the rest of the classifiers were used with their default settings. Additionally, Random Forest (RF) \cite{breiman2001random}, ADABOOST \cite{freund1997decision}, RUSBOOST \cite{seiffert2009rusboost}, SUBSPACE \cite{ho1998random}, TOTALBOOST \cite{warmuth2006totally}, and LPBOOST \cite{grove1998boosting} were employed with their default parameter values. Furthermore, the parameters for four novel ensemble classifier methods, namely PSOEL \cite{jan2020novel}, EAEL \cite{md2019evolutionary}, FESM \cite{jan2020anovel}, and GA-Bagging-SVM \cite{lin2019accurate}, were set to be consistent with the original papers. 

In our experiments, the original data was randomly divided into training and test datasets with an 8:2 ratio. The five-fold cross-validation method was used for training the data. For the GWO algorithm in feature selection and ensemble optimization, the population size ($P$) was set to 100, the number of iterations was set to 50, and the threshold ($\theta$) was set to 0.5. Specifically, the threshold value $\theta$ in our study is utilized as a criterion within the Grey Wolf Optimizer algorithm to determine feature selection, and is not directly related to actual gene expression values themselves. In the clustering phase, the parameter $t$ was set to $\sqrt[5]m$. The detailed parameters of seven classical evolutionary algorithms, including ACO \cite{dorigo2006ant}, CS \cite{yang2009cuckoo}, DE \cite{das2010differential}, GA \cite{whitley1994genetic}, GWO \cite{mirjalili2014grey}, PSO \cite{kennedy1995particle}, and ABC \cite{karaboga2007powerful}, are summarized in TABLE III, where the population size ($P$) and the maximum iteration ($max_t$) are set to the same values.

{
\begin{table}[!htb]
\centering
\caption{Parameters of Different Machine Learning Methods}
\resizebox{87mm}{!}{
\begin{tabular}{| l | l |}
\hline
\textbf{Methods} & \textbf{Parameters}  \\
\hline \hline
DISCR &  discrimtype = diaglinear\\
\hline
KNN & K = 3\\
\hline
SVM & 'KernelFunction' = 'rbf', 'IterationLimit' = 50000, 'Standardize'= true\\
\hline
NB & distribution = kernel \\
\hline
\end{tabular}}
\end{table}
}

\begin{table}[!htb]
\centering
\caption{Parameters of Different Evolutionary Algorithms}
\resizebox{87mm}{!}{
\begin{tabular}{| l | l |}
\hline
\textbf{Methods} & \textbf{Parameters}  \\
\hline \hline
ACO &  tau = 1, eta = 1, alpha = 1, beta = 0.1, rho = 0.2, Pop = 100 , $max_t$ = 50.\\
\hline
CS & lb = 0, ub = 1, $\theta = 0.5$, Pa = 0.25, alpha = 1, beta = 1.5, Pop = 100 , $max_t$ = 50.\\
\hline
DE & lb = 0, ub = 1, $\theta = 0.5$, CR = 0.9, F = 0.5, Pop = 100 , $max_t$ = 50.\\
\hline
GA & CR = 0.8, MR = 0.01, Pop = 100 , $max_t$ = 50.\\
\hline
PSO & lb = 0, ub = 1, $\theta =0.5$, c1 = 2, c2 = 2, w = 0.9, Vmax =(ub - lb)/2, Pop = 100 , $max_t$ = 50.\\
\hline
ABC & lb = 0, ub = 1, $\theta =0.5$, maxlimit = 5, Pop = 100 , $max_t$ = 50.\\
\hline
GWO & lb = 0, ub = 1, $\theta =0.5$, Pop = 100 , $max_t$ = 50.\\ 
\hline
\end{tabular}}
\end{table}

For ACO, "tau" denotes the pheromone value, "eta" denotes the heuristic desirability, "alpha" denotes the control pheromone, "beta" denotes the control heuristic, and "rho" denotes the pheromone trail decay coefficient, which is set to 0.2. For CS, "Pa" denotes the discovery rate, "alpha" denotes the constant, and "beta" denotes the Levy component. For DE, "CR" denotes the crossover rate, and "F" denotes the scale factor. For GA, "CR" denotes the crossover rate, and "MR" denotes the mutation rate. For PSO, "c1" denotes the cognitive factor, "c2" denotes the social factor, "w" denotes the inertia weight, and "Vmax" denotes the maximum velocity.

\section{Results and Analysis}
\subsection{Performance Comparisons with Other Nature-inspired Ensemble Learning Algorithms}
In our study, we conducted performance comparisons of EODE with several other nature-inspired ensemble learning algorithms, namely PSOEL, EAEL, FESM, and GA-Bagging-SVM. The experimental results are summarized in Figure 3, where Figure 3(A) presents detailed classification results, Figure 3(B) illustrates the performance comparisons of EODE against the other ensemble methods, and Figure 3(C) showcases the average performance values of these methods.

As shown in Figure 3(A), EODE achieved the best results among all methods on 26 out of the 35 datasets. Specifically, EODE attained 100\% classification accuracy on 7 datasets and achieved over 90\% accuracy on more than half of the datasets. These results highlight the robustness of EODE in handling various types of cancers and its ability to provide highly accurate classifications. From Figure 3(B), it is evident that EODE outperformed the other nature-inspired ensemble learning algorithms. The performance comparisons clearly demonstrate the superiority of EODE in terms of test accuracy. To provide a comprehensive performance overview, we present the average performance across all 35 cancer gene expression datasets in Figure 3(C). The results indicate that EODE outperformed PSOEL by 6\% and exhibited more than a 10\% improvement compared to the other methods. These findings strongly support the conclusion that EODE performs better than other nature-inspired ensemble methods in the context of cancer gene expression classification.

   
\begin{figure}[htb]
\centering
\includegraphics[scale = 0.33]{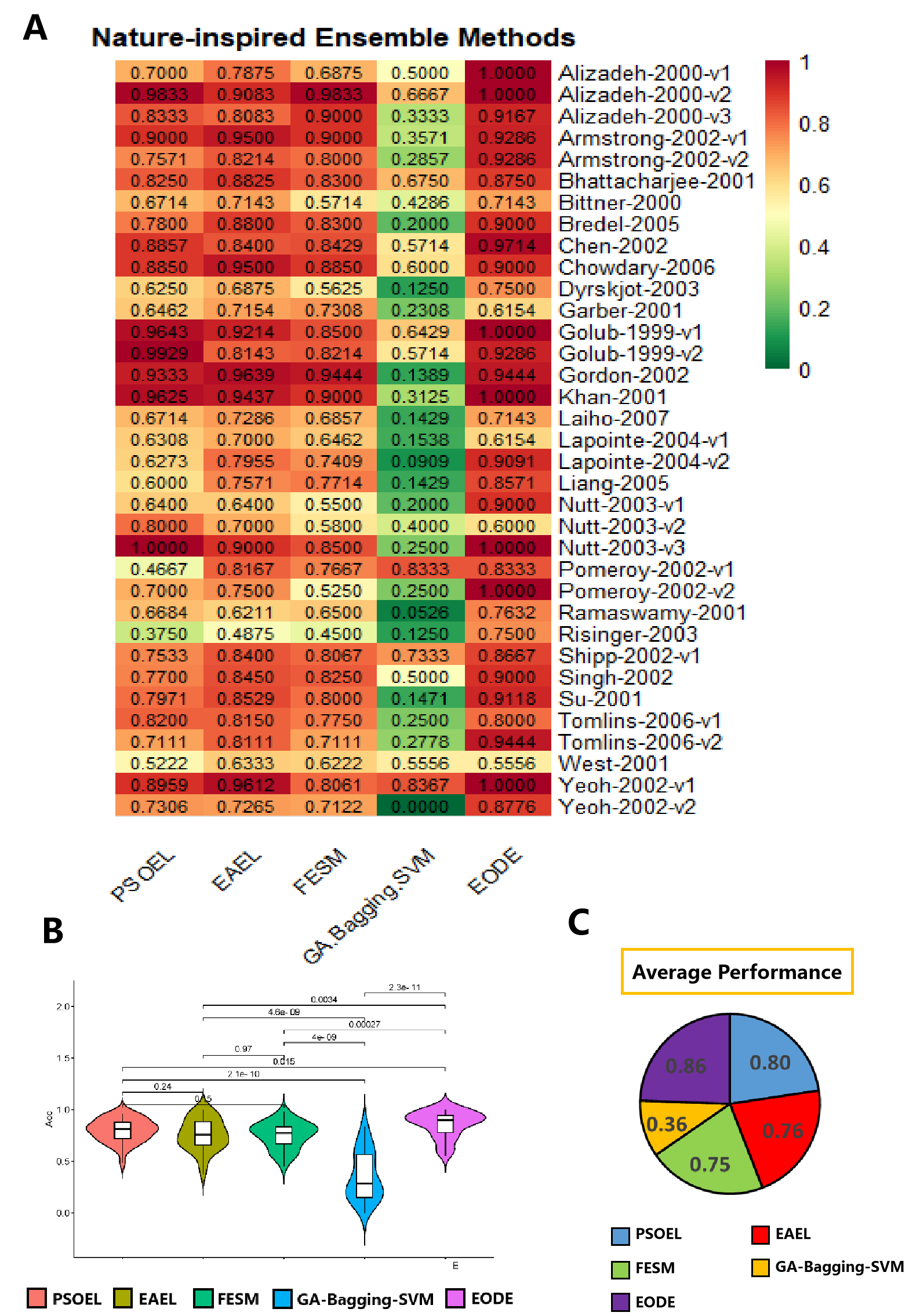}
\caption{Performance comparison to the other nature-inspired ensemble learning algorithms. \textit{(A)} Test classification results of EODE and four other nature-inspired ensemble methods across the 35 cancer gene expression datasets. \textit{(B)} Comparison graphs of EODE and the other four nature-inspired ensemble methods. \textit{(C)} The average performance of EODE and the other four nature-inspired ensemble methods across the 35 cancer gene expression datasets.}
\end{figure}

\subsection{Performance Comparisons of Different Machine Learning Algorithms}
In our study, we conducted a comprehensive analysis and comparison of the performance between our proposed ensemble approach, EODE, and single classifier approaches. The experimental results, as shown in Figure 4, clearly demonstrate the superiority of EODE in terms of classification accuracy for cancer gene expression datasets. EODE achieved the best classification accuracy for over 55\% of the datasets, surpassing all single classifiers. This indicates the effectiveness and robustness of our ensemble approach in handling cancer gene expression classification tasks. Moreover, when considering the average performance across all 35 cancer gene expression datasets, EODE consistently outperformed all single classifiers. Specifically, our ensemble approach exhibited remarkable improvements compared to the worst classifier, with an increase in performance of nearly 33\%. Furthermore, EODE consistently achieved performance improvements of more than 10\% compared to the majority of the base classifiers. 

These findings clearly highlight the advantages of our ensemble approach over traditional single classifier methods. By leveraging the collective wisdom of multiple classifiers, EODE effectively addresses the challenges posed by cancer gene expression classification, resulting in superior classification accuracy and overall performance. Figure 4 provides a visual representation of the experimental results, further supporting the conclusions drawn from our performance comparisons. The results validate the effectiveness of our proposed ensemble approach, highlighting its potential as a valuable tool in the field of cancer gene expression analysis.

\begin{figure}[htb!]
\centering
\includegraphics[scale = 0.26]{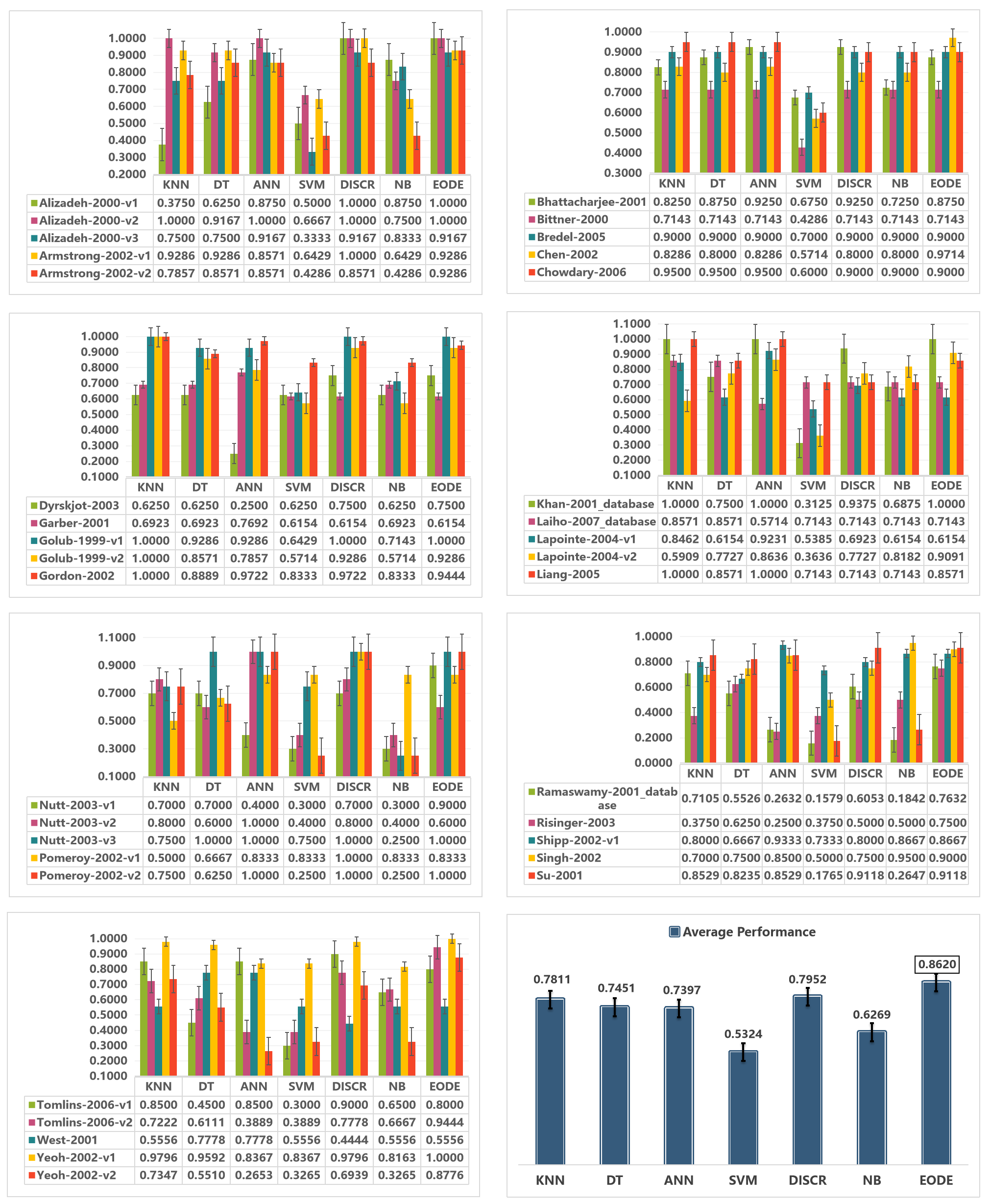}
\caption{Performance comparisons of the different machine learning algorithms. The first 7 graphs represent the test classification accuracy on the different cancer gene expression datasets, and the last graph indicates the average performance of the seven methods on the 35 datasets. }
\end{figure}

\subsection{Performance Comparisons of the Different Evolutionary Algorithms}
To further evaluate the performance of the proposed EODE method, we compared it against other state-of-the-art evolutionary algorithms, including: Ant Colony Optimization (ACO), Cuckoo Search (CS), Differential Evolution (DE), Genetic Algorithm (GA), Grey Wolf Optimizer (GWO), Particle Swarm Optimization (PSO), and Artificial Bee Colony (ABC).

The experimental results are summarized in supplementary Figures 1 and 2. Supplementary Figure 1 shows the classification accuracy of different methods on each of the 35 cancer gene expression datasets, with the first 7 sub-figures presenting results on individual datasets and the last sub-figure reporting the average performance across all datasets. As seen in supplementary Figure 1, EODE obtains the best classification accuracy on over 60\% of the datasets. Notably, there is an improvement of 5-8\% in average classification accuracy achieved by EODE compared to other evolutionary algorithms. Supplementary Figure 2 depicts the number of features (i.e. biomarker genes) selected by each method on each dataset. We can observe that EODE selects the smallest feature subset in nearly 60\% of datasets, indicating its ability to identify the most informative genes.

Overall, from both supplementary Figures 1 and 2, we can deduce that EODE consistently demonstrates the best average performance across all 35 cancer gene expression datasets, outperforming other state-of-the-art evolutionary methods. This validates the effectiveness and robustness of the proposed EODE approach in discovering critical biomarker genes for cancer classification.

\begin{table*}[!htb]
\centering
\caption{Performance on Training and Testing Sets with and without Ensemble Learning (WEL) Method.}
\resizebox{170mm}{!}{
\begin{tabular}{| l | l | l | l | l | l | l | l | l | l |}
\hline
\multicolumn{1}{|l|}{\multirow{2}{*}{Datasets}} & \multicolumn{2}{c|}{Training Accuracy}  & \multicolumn{2}{c|}{Test Accuracy} & \multicolumn{1}{l|}{\multirow{2}{*}{Datasets}} & \multicolumn{2}{c|}{Training Accuracy} & \multicolumn{2}{c|}{Test Accuracy}   \\ \cline{2-5} \cline{7-10} 
\multicolumn{1}{|l|}{}   & \multicolumn{1}{l|}{WEL} & \multicolumn{1}{l|}{EODE} & \multicolumn{1}{l|}{WEL} & \multicolumn{1}{l|}{EODE} & \multicolumn{1}{l|}{}  & \multicolumn{1}{l|}{WEL} & \multicolumn{1}{l|}{EODE} & \multicolumn{1}{l|}{WEL} & \multicolumn{1}{l|}{EODE} \\ 
\hline \hline
Alizadeh-2000-v1	&	\textbf{1.0000}	&	\textbf{1.0000}	&	0.5114	&	\textbf{1.0000}	&	Lapointe-2004-v2	&	\textbf{0.8737}	&	0.8634	&	0.4339	&	\textbf{0.9091}	\\	\hline
Alizadeh-2000-v2	&	\textbf{1.0000}	&	\textbf{1.0000}	&	0.6591	&	\textbf{1.0000}	&	Liang-2005	&	0.9515	&	\textbf{0.9667}	&	0.6494	&	\textbf{0.8571}	\\	\hline
Alizadeh-2000-v3	&	\textbf{0.9745}	&	0.9600	&	0.4318	&	\textbf{0.9167}	&	Nutt-2003-v1	&	0.8614	&	\textbf{0.9250}	&	0.4273	&	\textbf{0.9000}	\\	\hline
Armstrong-2002-v1	&	\textbf{1.0000}	&	0.9818	&	0.6558	&	\textbf{0.9286}	&	Nutt-2003-v2	&	\textbf{1.0000}	&	\textbf{1.0000}	&	0.5455	&	\textbf{0.6000}	\\	\hline
Armstrong-2002-v2	&	\textbf{1.0000}	&	\textbf{1.0000}	&	0.4416	&	\textbf{0.9286}	&	Nutt-2003-v3	&	\textbf{1.0000}	&	\textbf{1.0000}	&	0.7955	&	\textbf{1.0000}	\\	\hline
Bhattacharjee-2001	&	\textbf{0.9944}	&	0.9938	&	0.6750	&	\textbf{0.8750}	&	Pomeroy-2002-v1	&	\textbf{0.9697}	&	0.9667	&	\textbf{0.8333}	&	\textbf{0.8333}	\\	\hline
Bittner-2000	&	0.9614	&	\textbf{0.9667}	&	0.5857	&	\textbf{0.7143}	&	Pomeroy-2002-v2	&	\textbf{0.9532}	&	0.9381	&	0.2841	&	\textbf{1.0000}	\\	\hline
Bredel-2005	&	0.9114	&	\textbf{0.9250}	&	0.5727	&	\textbf{0.9000}	&	Ramaswamy-2001$\_$database	&	0.7251	&	\textbf{0.7966}	&	0.1794	&	\textbf{0.7632}	\\	\hline
Chen-2002	&	0.9918	&	\textbf{0.9931}	&	0.6649	&	\textbf{0.9714}	&	Risinger-2003	&	\textbf{0.8649}	&	0.8238	&	0.4545	&	\textbf{0.7500}	\\	\hline
Chowdary-2006	&	\textbf{0.9904}	&	0.9875	&	0.8682	&	\textbf{0.9000}	&	Shipp-2002-v1	&	0.9800	&	\textbf{0.9833}	&	0.7212	&	\textbf{0.8667}	\\	\hline
Dyrskjot-2003	&	\textbf{0.9948}	&	0.9381	&	0.5568	&	\textbf{0.7500}	&	Singh-2002	&	\textbf{0.9778}	&	0.9750	&	0.5955	&	\textbf{0.9000}	\\	\hline
Garber-2001	&	\textbf{0.8883}	&	0.8873	&	0.5455	&	\textbf{0.6154}	&	Su-2001	&	0.9565	&	\textbf{0.9643}	&	0.1925	&	\textbf{0.9118}	\\	\hline
Golub-1999-v1	&	0.9970	&	\textbf{1.0000}	&	0.6234	&	\textbf{1.0000}	&	Tomlins-2006-v1	&	0.8787	&	\textbf{0.8809}	&	0.3364	&	\textbf{0.8000}	\\	\hline
Golub-1999-v2	&	0.9939	&	\textbf{1.0000}	&	0.5065	&	\textbf{0.9286}	&	Tomlins-2006-v2	&	\textbf{0.8721}	&	0.8648	&	0.3889	&	\textbf{0.9444}	\\	\hline
Gordon-2002	&	\textbf{1.0000}	&	\textbf{1.0000}	&	0.8409	&	\textbf{0.9444}	&	West-2001	&	\textbf{1.0000}	&	\textbf{1.0000}	&	0.4646	&	\textbf{0.5556}	\\	\hline
Khan-2001$\_$database	&	\textbf{1.0000}	&	\textbf{1.0000}	&	0.3466	&	\textbf{1.0000}	&	Yeoh-2002-v1	&	\textbf{0.9950}	&	\textbf{0.9950}	&	0.8108	&	\textbf{1.0000}	\\	\hline
Laiho-2007$\_$database	&	\textbf{1.0000}	&	\textbf{1.0000}	&	\textbf{0.7143}	&	\textbf{0.7143}	&	Yeoh-2002-v2	&	0.8448	&	\textbf{0.8995}	&	0.2430	&	\textbf{0.8776}	\\	\hline
Lapointe-2004-v1	&	0.8423	&	\textbf{0.8591}	&	0.5385	&	\textbf{0.6154}	&	\textbf{Average}	&	0.9498	&	\textbf{0.9524}	&	0.5456	&	\textbf{0.8620}	\\	\hline
\end{tabular}}
\end{table*}

\subsection{Performance Comparisons of the Different Ensemble Learning Algorithms}
\begin{figure}[htb!]
\centering
\includegraphics[scale = 0.22]{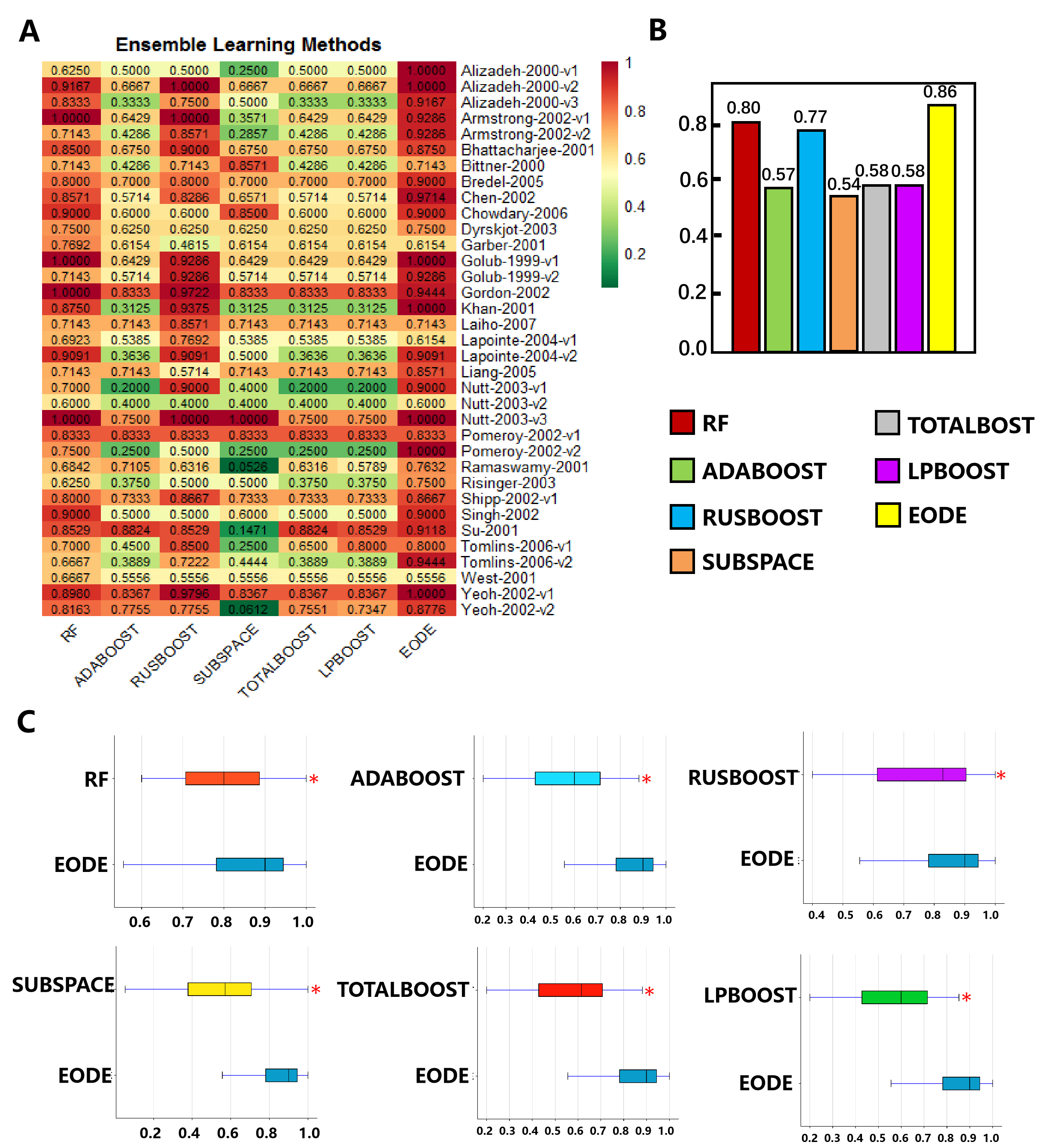}
\caption{Performance comparisons of the different ensemble learning algorithms. \textit{(A)} Test classification results of EODE and six other ensemble methods across the 35 cancer gene expression datasets;  \textit{(B)} The average performance of EODE and the six other ensemble classifiers on the 35 datasets; \textit{(C)} Graphs of EODE versus the other ensemble classifiers, where RF denotes Random Forest. }
\end{figure}

To further validate the effectiveness of the proposed EODE method, we conducted experiments comparing its performance to other state-of-the-art ensemble learning classifiers on the 35 cancer gene expression datasets. The methods considered for comparison include: Random Forest (RF) \cite{breiman2001random}, ADABOOST \cite{freund1997decision}, RUSBOOST \cite{seiffert2009rusboost}, SUBSPACE \cite{ho1998random}, TOTALBOOST \cite{warmuth2006totally} and LPBOOST \cite{grove1998boosting}. 

The results are shown in Figure 5. Figure 5(A) depicts a heat map of the classification accuracy of different methods on each of the 35 datasets, where darker colors indicate better performance. This heat map visualization allows us to qualitatively compare the performance of models across various cancer types. Figure 5(B) summarizes the mean classification accuracy of each method averaged over the 35 datasets. The proposed EODE method achieves 6-32\% better performance compared to other ensemble learning classifiers, demonstrating its superior predictive ability. Figure 5(C) presents box plots to compare the distribution of classification accuracies obtained by each method on different datasets. We can observe that the median accuracy of EODE is higher than all other methods, indicating its stable and robust performance. Moreover, the box plot of EODE is more narrow compared to others, showing the consistency of results obtained.

Overall, these quantitative and qualitative comparisons presented in Figure 5 validate that the proposed EODE method achieves the best classification accuracy on over 70\% of the cancer gene expression datasets, outperforming other state-of-the-art ensemble classifiers. This clearly demonstrates the effectiveness and robustness of the EODE approach for cancer classification using gene expression data.

\subsection{Ablation Study}
\subsubsection{Performance of EODE without Ensemble Learning}
TABLE 4 presents a comprehensive evaluation of training and testing performance across 35 datasets using the proposed EODE approach against EODE without ensemble learning (WEL).
Here, WEL means the nature-inspired diverse ensemble learning is not employed. On analysis, it is evident that the training accuracy of both EODE and WEL are comparable, with the EODE approach achieving a slightly higher average training accuracy of 0.9524 versus 0.9498 for WEL. This indicates that the model capacity for fitting the training data is similar between the two approaches. However, EODE demonstrates a significant test accuracy advantage over WEL, with average test accuracies of 0.8620 and 0.5456 respectively. This translates to an absolute improvement of over 30\% in generalization performance by leveraging ensemble learning.

The key insight is that while ensemble learning does not markedly improve training fit, it provides superior generalization through effectively preventing overfitting. Single models are prone to overfitting the noise in small datasets. Ensemble learning creates multiple diverse models and aggregates their predictions, avoiding these spurious patterns. Across multiple datasets, EODE consistently exhibits stronger generalization, evidenced by the significantly higher test accuracies. This gap is particularly prominent in smaller datasets where individual models tend to overfit more. By reducing variance via ensembling, the proposed approach demonstrates more robust predictions on unseen test data. The results validate the effectiveness of ensemble learning in enhancing model generalization capability and tackling the overfitting challenge.

In conclusion, the ensemble framework shows considerable promise in boosting test performance over single model baseline across a wide range of conditions. This has important implications for real-world applications like speech emotion recognition where avoiding overfitting is critical. The analysis provides strong empirical evidence and rationale for adopting ensemble techniques. 

\subsubsection{ Performance of EODE Ensemble versus Individual Classifiers}
Unlike the analysis in Section 4.2, this study does not evaluate each base classifier model in isolation. Rather, this section investigates the impact of using a single base classifier within the nature-inspired diverse ensemble learning phase of the proposed approach, instead of aggregating multiple heterogeneous classifiers concurrently as intended in the ensemble methodology. By focusing on the ensemble learning stage, this analysis provides targeted insight into the benefits of leveraging diversity in the classifier combinations compared to relying on any individual modeling paradigm alone during this critical step. 

\begin{table}[!htb]
\centering
\caption{Performance of EODE Ensemble versus Individual Base Classifiers}
\resizebox{90mm}{!}{
\begin{tabular}{| l | l | l | l | l | l | l | l | }
\hline
\textbf{Dataset} & \textbf{DISCR} & \textbf{DT}& \textbf{KNN} & \textbf{ANN} & \textbf{SVM} & \textbf{NB} & \textbf{EODE} \\
\hline \hline
Alizadeh-2000-v1	&	0.8333	&	0.6875	&	0.7750	&	0.6875	&	0.5000	&	0.7500	&	\textbf{1.0000}	\\	\hline
Alizadeh-2000-v2	&	\textbf{1.0000}	&	0.8452	&	0.9833	&	\textbf{1.0000}	&	0.6667	&	0.7333	&	\textbf{1.0000}	\\	\hline
Alizadeh-2000-v3	&	\textbf{0.9306}	&	0.7381	&	0.8667	&	0.9271	&	0.3333	&	0.8000	&	0.9167	\\	\hline
Armstrong-2002-v1	&	0.8929	&	\textbf{0.9490}	&	0.9143	&	0.8929	&	0.6429	&	0.8429	&	0.9286	\\	\hline
Armstrong-2002-v2	&	0.8690	&	0.7959	&	0.8000	&	0.8482	&	0.4286	&	0.6714	&	\textbf{0.9286}	\\	\hline
Bhattacharjee-2001	&	\textbf{0.9292}	&	0.8321	&	0.8350	&	0.9063	&	0.6750	&	0.7750	&	0.8750	\\	\hline
Bittner-2000	&	0.6905	&	0.6531	&	\textbf{0.7143}	&	0.6786	&	0.4286	&	0.5714	&	\textbf{0.7143}	\\	\hline
Bredel-2005	&	0.8667	&	0.6857	&	0.7800	&	0.8500	&	0.7000	&	0.8400	&	\textbf{0.9000}	\\	\hline
Chen-2002	&	0.8619	&	0.7837	&	0.8171	&	0.8786	&	0.5829	&	0.8229	&	\textbf{0.9714}	\\	\hline
Chowdary-2006	&	0.9000	&	0.9214	&	0.9300	&	\textbf{0.9438}	&	0.8900	&	0.8700	&	0.9000	\\	\hline
Dyrskjot-2003	&	\textbf{0.7500}	&	0.6429	&	0.6750	&	0.6719	&	0.6250	&	0.6500	&	\textbf{0.7500}	\\	\hline
Garber-2001	&	\textbf{0.7051}	&	0.6923	&	0.6769	&	0.6538	&	0.6154	&	0.6769	&	0.6154	\\	\hline
Golub-1999-v1	&	0.9286	&	0.9184	&	0.9143	&	0.8482	&	0.6429	&	0.8000	&	\textbf{1.0000}	\\	\hline
Golub-1999-v2	&	0.8810	&	0.8367	&	0.8857	&	0.7500	&	0.5714	&	0.6714	&	\textbf{0.9286}	\\	\hline
Gordon-2002	&	\textbf{0.9861}	&	0.9643	&	0.9722	&	0.9757	&	0.8333	&	0.9500	&	0.9444	\\	\hline
Khan-2001$\_$database	&	0.8646	&	0.8571	&	0.9125	&	0.9297	&	0.3125	&	0.7000	&	\textbf{1.0000}	\\	\hline
Laiho-2007$\_$database	&	0.7381	&	\textbf{0.7551}	&	0.7143	&	0.6786	&	0.7143	&	0.7429	&	0.7143	\\	\hline
Lapointe-2004-v1	&	0.7538	&	0.6264	&	\textbf{0.8154}	&	0.6923	&	0.5385	&	0.6000	&	0.6154	\\	\hline
Lapointe-2004-v2	&	0.7386	&	0.5195	&	0.5909	&	0.7330	&	0.3636	&	0.7091	&	\textbf{0.9091}	\\	\hline
Liang-2005	&	0.6786	&	0.7347	&	0.8000	&	\textbf{0.8750}	&	0.7143	&	0.7143	&	0.8571	\\	\hline
Nutt-2003-v1	&	0.7000	&	0.3857	&	0.5800	&	0.5500	&	0.2800	&	0.3800	&	\textbf{0.9000}	\\	\hline
Nutt-2003-v2	&	\textbf{0.8000}	&	0.4286	&	0.6800	&	0.6500	&	0.4400	&	0.4000	&	0.6000	\\	\hline
Nutt-2003-v3	&	\textbf{1.0000}	&	0.8214	&	0.8500	&	\textbf{1.0000}	&	0.7500	&	0.8500	&	\textbf{1.0000}	\\	\hline
Pomeroy-2002-v1	&	0.6250	&	0.6429	&	0.6333	&	0.6250	&	0.1667	&	0.6000	&	\textbf{0.8333}	\\	\hline
Pomeroy-2002-v2	&	0.8750	&	0.4107	&	0.5500	&	0.5156	&	0.2500	&	0.2500	&	\textbf{1.0000}	\\	\hline
Ramaswamy-2001	&	0.6250	&	0.5752	&	0.6263	&	0.5329	&	0.1579	&	0.2526	&	\textbf{0.7632}	\\	\hline
Risinger-2003	&	0.5938	&	0.5000	&	0.3750	&	0.5625	&	0.3750	&	0.5000	&	\textbf{0.7500}	\\	\hline
Shipp-2002-v1	&	0.7667	&	0.7619	&	0.8000	&	\textbf{0.8667}	&	0.7333	&	0.7733	&	\textbf{0.8667}	\\	\hline
Singh-2002	&	0.8125	&	0.7833	&	0.7600	&	0.8188	&	0.5000	&	0.8500	&	\textbf{0.9000}	\\	\hline
Su-2001	&	0.8676	&	0.7549	&	0.7588	&	0.8750	&	0.1588	&	0.4471	&	\textbf{0.9118}	\\	\hline
Tomlins-2006-v1	&	0.7875	&	0.5250	&	\textbf{0.8200}	&	0.8000	&	0.3000	&	0.6800	&	0.8000	\\	\hline
Tomlins-2006-v2	&	0.7778	&	0.6000	&	0.7000	&	0.7986	&	0.3889	&	0.6556	&	\textbf{0.9444}	\\	\hline
West-2001	&	0.5833	&	\textbf{0.7778}	&	0.6444	&	0.5972	&	0.4667	&	0.6000	&	0.5556	\\	\hline
Yeoh-2002-v1	&	0.9796	&	0.9714	&	0.9878	&	0.9566	&	0.6694	&	0.8327	&	\textbf{1.0000}	\\	\hline
Yeoh-2002-v2	&	0.6735	&	0.6408	&	0.7673	&	0.5357	&	0.3265	&	0.3347	&	\textbf{0.8776}	\\	\hline
\textbf{Average}	&	0.8076	&	0.7148	&	0.7687	&	0.7744	&	0.5069	&	0.6656	&	\textbf{0.8620}	\\	\hline
\end{tabular}}
\end{table}

Across the 35 gene expression datasets analyzed, the best single classifier achieved an average accuracy of 0.8076 using a DISCR model. In contrast, the proposed EODE ensemble approach attained a significantly higher accuracy of 0.8620 by leveraging an integrated combination of diverse classifiers including DISCR, DT, KNN, ANN, SVM and NB. The results highlight that relying on any individual base classifier is suboptimal compared to the ensemble approach. No single modeling paradigm consistently dominates the performance across all datasets, due to the complexity of the classification problem. Different datasets exhibit variability in terms of which individual classifier achieves the best performance when used alone. However, EODE provides equal or higher accuracy relative to the top stand-alone model on 23 out of 35 datasets. The results empirically demonstrate that integrating multiple complementary base classifiers simultaneously is essential to maximize the potential of the ensemble framework and attain optimal classification performance on gene expression data. Reliance on any single constituent classifier within the ensemble learning process fails to harness the full synergistic advantages of the diverse ensemble.

\section{Conclusion} 
Cancer type identification is a critical aspect of cancer research, as it enables early diagnosis and tailored treatment for patients. One key challenge in this field is identifying the highly sensitive biomarker genes that are indicative of specific cancer types. In this study, we propose a novel approach called EODE to address the classification of cancer types, particularly in scenarios where the gene expression profile samples are high-dimensional and small in size. EODE leverages the grey wolf optimizer (GWO) to optimize feature subsets and collaboratively builds an optimized ensemble classifier. By combining nature-inspired feature selection and ensemble learning, EODE significantly improves the model's identification capability.

We conducted experiments on 35 datasets encompassing various cancer types, and the results demonstrate the effectiveness of our algorithm compared to four nature-inspired ensemble methods (PSOEL, EAEL, FESM, and GA-Bagging-SVM), six benchmark machine learning algorithms (KNN, DT, ANN, SVM, DISCR, and NB), six state-of-the-art ensemble algorithms (RF, ADABOOST, RUSBOOST, SUBSPACE, TOTALBOOST, and LPBOOST), and seven nature-inspired methods (ACO, CS, DE, GA, GWO, PSO, and ABC). Our algorithm outperformed these methods in terms of classification accuracy.

In future work, we aim to enhance the efficiency of the algorithm by improving the screening of redundant and invalid features. Additionally, as biomedical data often exhibit class imbalance, we plan to ensure robust results on class-imbalanced data. Beyond computational refinements, we intend to evaluate the proposed methodology on expanded gene expression datasets from diverse clinical cohorts. As cancer subtyping using gene expression data holds great promise for guiding individualized treatment decisions, we hope to transition this computational pipeline into real-world clinical settings.


%

\ifCLASSOPTIONcompsoc
  \section*{Acknowledgments}

\else
  \section*{Acknowledgment}
\fi

The work described in this paper was substantially supported by the National Natural Science
Foundation of China under Grant No. 62076109, and funded by the Natural Science Foundation of
Jilin Province under Grant No. 20190103006JH, the Natural Science Funds of Jilin Province under Grant No. 20200201158JC. The work described in this paper was supported by
the grant from the Health and Medical Research Fund, the Food and Health Bureau, The Government
of the Hong Kong Special Administrative Region [07181426], and the funding from Hong Kong
Institute for Data Science (HKIDS) at City University of Hong Kong. The work described in this
paper was partially supported by two grants from City University of Hong Kong (CityU 11202219,
CityU 11203520). This research is also supported by the National Natural Science Foundation of China under Grant No. 32000464.

\ifCLASSOPTIONcaptionsoff
  \newpage
\fi



%

\bibliographystyle{unsrt}
\bibliography{ref.bib}
%




\end{document}